\pdfoutput=1

\documentclass[11pt]{article}
\usepackage[table]{xcolor}

\usepackage{ACL2023}
\usepackage[hang,flushmargin]{footmisc}
\usepackage{balance}
\usepackage{changepage}

\usepackage{times}
\usepackage{latexsym}

\usepackage[T1]{fontenc}
\usepackage[utf8]{inputenc}

\usepackage{microtype}

\usepackage{inconsolata}

\usepackage{float}

\usepackage{amsmath}
\usepackage{amssymb}
\usepackage{amsthm}
\usepackage{bm}
\usepackage{booktabs}
\usepackage{graphicx}
\usepackage{multicol}
\usepackage{multirow}
\usepackage{blindtext}
\usepackage{dirtree}
\usepackage{tikz}

\usepackage{caption}
\usepackage{subcaption}
\usepackage{enumitem}
\usepackage{tipa}
\usepackage{pmboxdraw}
\usepackage{wasysym}

\newcommand{\parheader}[1]{{\bf \smallskip \noindent #1.}}
\newcommand{\parheaderfirst}[1]{{\noindent \bf #1.}}

\newcommand{\DONE}[1]{\noindent \textcolor{green}{\textbf{DONE}}\\ }

\definecolor{severe}{RGB}{241,113,31}
\definecolor{severe-moderate}{RGB}{215,75,63}
\definecolor{moderate}{RGB}{177,50,90}
\definecolor{moderate-light}{RGB}{135,33,107}
\definecolor{light}{RGB}{92,18,110}

\newcommand{\leaf}{\hspace{0.31mm}\raisebox{0.15mm}[0pt][0pt]{└}\hspace{0.1mm}}
\newcommand{\branch}{\hspace{0.31mm}\raisebox{0.15mm}[0pt][0pt]{├}\hspace{0.1mm}}

\title{Drawing Conclusions from Draws:\ Rethinking Draw\\ Semantics in Arena-Style LLM Evaluation}

\newcommand{\ignore}[1]{}

\author{Raphael Tang$^{1}$ ~Crystina Zhang$^2$ ~Wenyan Li$^3$ ~Carmen Lai$^4$ ~Pontus Stenetorp$^{1,5}$ ~Yao Lu$^1$\vspace{1mm}\\
$^1$Centre for Artificial Intelligence, University College London\\$^2$University of Waterloo~~~$^3$University of Copenhagen~~~$^4$Independent Researcher\\$^5$Research and Development Center for Large Language Models, National Institute of Informatics}
\begin{document}
\maketitle

\begin{abstract}
In arena-style evaluation of large language models (LLMs), two LLMs respond to a user query, and the user chooses the winning response or deems the ``battle'' a draw, resulting in an adjustment to the ratings of both models.
The prevailing approach for modeling these rating dynamics is to view battles as two-player game matches, as in chess, and apply the Elo rating system and its derivatives.
In this paper, we critically examine this paradigm.
Specifically, we question whether a draw genuinely means that the two models are equal and hence whether their ratings should be equalized.
Instead, we conjecture that draws are more indicative of query difficulty:\ if the query is too easy, then both models are more likely to succeed equally.
On three real-world arena datasets, we show that ignoring rating updates for draws yields a 1--3\% relative increase in battle outcome prediction accuracy (which \textit{includes} draws) for all four rating systems studied.
Further analyses suggest that draws occur more for queries rated as very easy and those as highly objective, with risk ratios of 1.37 and 1.35, respectively.
We recommend future rating systems to reconsider existing draw semantics and to account for query properties in rating updates.
\end{abstract}

\section{Introduction}
In arena-style evaluation, as popularized by Chatbot Arena~\cite{chiang2024chatbot}, users issue arbitrary queries to two large language models (LLMs) and judge their responses, either choosing the winner or declaring the ``battle'' a draw.
The battles are treated as two-player zero-sum games like chess, where wins, losses, and draws respectively indicate outperformance, underperformance, and equal ability.
Naturally, most applications model these rating dynamics using the Elo rating system~\cite{elo1978rating} and its numerous derivatives~\cite{glickman2024models}:\ wins increase a model's rating at the expense of the opposing model, and draws equalize the ratings of the two models.

In this paper, we critically examine this two-player game paradigm, specifically questioning whether draws genuinely mean skill parity and thus whether ratings should be equalized.
Instead, we conjecture that draws mostly predict query difficulty and subjectivity.
If the query is too easy, both models are more likely to succeed equally.
Likewise, if the query is objective as opposed to subjective, the likelihood for both models to arrive at the same answer increases as well.
In short, we hypothesize that draws relate more strongly to query properties rather than equality of model ability.

We validate our hypothesis with two main experiments:\ First, we evaluate the effectiveness of established rating systems when rating updates for draws are ignored.
If draws are uninformative, there should be no difference in rating quality.
Across four rating systems and three real-world datasets, we find that ignoring draw updates \textit{increases} battle prediction accuracy by a relative 1--3\%, despite the evaluation \textit{still including} draws.
The improvement was consistently present for 11 of the 12 dataset--rating system combinations.
Second, we investigate how query difficulty, subjectivity, and model ratings relate to the probability of observing a draw.
We show that queries with difficulty and subjectivity ratings of 0 out of 5 are associated with a 35--37\% increased relative risk of observing a draw, whereas rating proximity has no substantial connection to draw probability.

In summary, our main contributions are (1) to our knowledge, we are the first to demonstrate that draws largely do not indicate model parity in arena-style evaluation, and (2) we provide insights into draw semantics, finding that query difficulty and subjectivity are better predictors of draw likelihood than model rating closeness.
Our work suggests the reconsideration of draw semantics in arena-style evaluation and the inclusion of query properties in rating updates.
We release our codebase at \url{https://github.com/daemon/lmarena-draws}.

\section{Arena-Style Evaluation}
\subsection{Preliminaries}
Arena-style evaluation comprises two stages:\ user judgement elicitation and model rating updates.
First, users interact with a pair of anonymous LLMs and provide judgements, either picking the better response or declaring them to be the same---see Figure~\ref{fig:example-ui} in Appendix~\ref{sec:appendix-ui} for an example user interface.
Next, the system updates the two models' ratings based on the judgement, with the winning model receiving points at the expense of the losing model.
If the battle is a draw, then the rating system equalizes the two ratings, subtracting from the higher rating and adding to the lower one.

Formally, let $M$ be a finite set of models and $L$ be the set $\{1, 0, \frac{1}{2}\}$ denoting win, loss, and draw, respectively.
Then let $\big((m_{a_i}, m_{b_i}, y_i)\big)_{i=1}^n$ be the time-ordered sequence of $n$ battles, where $m_{a_i}, m_{b_i} \in M$ denote the two models and $y_i \in L$ is the judgement of $m_{a_i}$ relative to $m_{b_i}$, e.g., $1$ means $m_{a_i}$ won against $m_{b_i}$.
The rating system initializes each model $m\in M$ indexed by $j \in \mathbb{N}$ with a rating $r^{(j)}_1 \in \mathbb{R}$, which is updated after each battle by the system's update rule\vspace{-1mm}
\begin{equation}
\small
\big(r^{\left({a_i}\right)}_{i+1}, r^{\left({b_i}\right)}_{i+1}\big) := f\Big(r^{\left({a_i}\right)}_i, r^{\left({b_i}\right)}_i, y_i\Big),\vspace{-1mm}
\end{equation}
where $f : \mathbb{R} \times \mathbb{R} \times L \mapsto \mathbb{R} \times \mathbb{R}$ takes two model ratings at timestep $i$ and the battle outcome $y_i$ to produce two updated ratings for the next timestep.
At timestep $i+1$ for all $r^{(j)}_i$ where $j$ is neither ${a_i}$ nor ${b_i}$, the rating is unchanged, i.e., $r^{(j)}_{i+1} := r^{(j)}_{i}$.

\subsection{Rating Systems}
Online score-based rating systems primarily vary in their update rules $f$.
In this paper, we consider four established rating systems: Elo~\cite{elo1978rating}, popular in competitive chess; Glicko-2~\cite{glickman2012example}, an alternative model for chess; online Bradley--Terry, as implemented by Chatbot Arena~\cite{chiang2023chatbot}; and TrueSkill~\cite{herbrich2006trueskill}, a Bayesian system originally designed for matchmaking on Xbox Live.

\parheader{Elo}
Elo proposes a logistic model for the expected probabilities $E_{{a_i}}$ of $m_{a_i}$ or $m_{b_i}$ winning:\vspace{-2.5mm}
\begin{equation}\small
    E_{{a_i}} := 1/\Big(1 + 10^\frac{r_i^{(m_{b_i})} - r_i^{(m_{a_i})}}{400} \Big),~~E_{{b_i}} := 1 - E_{{a_i}},\vspace{-1mm}
\end{equation}
then uses an update rule with learning rate $K$ (the $K$-factor) given the actual observed outcome:\vspace{-2.5mm}
\begin{equation}\footnotesize
    r^{({a_i})}_{i+1} := r_i^{(m_{a_i})} + K(y_i - E_{{a_i}}),
\end{equation}\vspace{-7mm}
\begin{equation}\footnotesize
    r^{({b_i})}_{i+1} := r_i^{(m_{b_i})} + K((1 - y_i) - E_{{b_i}}).
\end{equation}

\parheaderfirst{Glicko-2}
\citet{glickman2012example} is another logistic model that additionally tracks the rating deviation~(RD) $\phi_i^{(j)}$ and volatility $\sigma^{(j)}_i$.
Let the weighting function be $g(\phi_i^{(j)}) := (1+3{\phi_i^{(j)}}^2/\pi^2)^{-\frac{1}{2}}$ and the expected win probability
\begin{equation}\small
E_{a_i} := 1/\big(1+\exp(-g(\phi_i^{(b_i)})(r^{(a_i)}_i - r^{(b_i)}_i))\big).
\end{equation}
Let the variance $v_i:= \big(g(\phi_i^{(b_i)})^2E_{a_i} (1 - E_{a_i})\big)^{-1}$ and the delta be $\Delta:=v_ig(\phi_i^{(b_i)})(y_i-E_{a_i})$.
The update rule for $\sigma_{i+1}^{(a_i)}$ is then the root of
\begin{equation}\small
    h(x) := \frac{e^x(\Delta^2-\phi_i^{(a_i)}-v_i-e^x)}{2(\phi^{(a_i)}_i + v_i + e^x)^2}-\frac{x-2\ln\sigma_{i}^{(a_i)}}{\tau^2},
\end{equation}
where $\tau >0$ is a constant for volatility change.
The RD is updated as $\phi_{i+1}^{(a_i)} := \big(1/({\phi_i^{(a_i)}}^2+\sigma^{(a_i)}_{i+1})+1/v\big)^{-\frac{1}{2}}$ and the rating as $r^{(a_i)}_{i+1} := r_i^{(a_i)} + {\phi^{(a_i)}_{i+1}}^2 g(\phi_i^{(b_i)})(y_i-E_{a_i})$.
The update rules for $m_{b_i}$ proceed analogously with $y_i \mapsto 1-y_i$.
Intuitively, Glicko-2 scales the size of its updates with the level of uncertainty and volatility.

\parheader{Bradley--Terry}
Chatbot Arena adopts an online Bradley--Terry model~\cite{bradley1952rank} over Elo due to its greater stability~\cite{chiang2023chatbot}.
The update rule for $y_i \in \{0, 1\}$ is
\begin{equation}
    r^{(a_i)}_{i+1} :=  r_i^{(a_i)} + \eta (y_i - E_{a_i}),
\end{equation}
where $\eta > 0$ is the learning rate and $E_{a_i} := 1/\big(1+\exp(r_i^{(b_i)} - r_i^{(a_i)})\big)$ is the probability of $m_{a_i}$ winning against $m_{b_i}$.
The other model rating $r_i^{(b_i)}$ is updated similarly with $y_i$ flipped.
For draws, i.e., $y_i = 1/2$, Chatbot Arena~\cite{chiang2023chatbot} performs a simultaneous win and a loss update, effectively reducing the gap between the two ratings.

\parheader{TrueSkill}
\citet{herbrich2006trueskill} introduce a Bayesian rating system that treats ratings as Gaussian priors $\mathcal{N}(r_i^{(j)}, {\sigma_i^{(j)}}^2)$ in a factor graph.
Each battle draws a performance $p_i^{(j)}\sim \mathcal{N}(s^{(j)}_i, \beta^2)$, where $s^{(j)}_i \sim\mathcal{N}(r_i^{(j)}, {\sigma_i^{(j)}}^2)$, and the probability of $m_{a_i}$ winning against $m_{b_i}$ is modeled as the truncated Gaussian over the performance difference:
\begin{equation}\scriptsize
    E_{a_i} := 1 - \Phi\Bigg(\frac{\varepsilon-(r_i^{(a_i)} - r_i^{(b_i)})}{\sqrt{2\beta^2+{\sigma_i^{(a_i)}}^2+{\sigma_i^{(b_i)}}^2}}\Bigg),
\end{equation}
where $\Phi$ is the Gaussian CDF and $\varepsilon >0$ is the draw margin.
The hard evidence (likelihood) is the outcome $y_i$, and the new rating posterior ($r^{(a_i)}_{i+1}$, ${\sigma_{i+1}^{(a_i)}}^2$) is computed using a full Bayesian update with message passing over the factor graph; see \citet{herbrich2006trueskill} for closed-form equations.

\begin{table*}[t]
\setlength{\tabcolsep}{2.5pt}
\centering
\small
    \begin{tabular}{lllllllc}
    \toprule[1pt]
    \multirow{2}{*}{\vspace{-1ex}Method} & \multicolumn{2}{c}{LMArena} & \multicolumn{2}{c}{SearchArena} & \multicolumn{2}{c}{VisionArena} & \multirow{2}{*}{\vspace{-1ex}$\Delta$\%}\\
    \cmidrule(lr){2-3} \cmidrule(lr){4-5} \cmidrule(lr){6-7} &  Acc. & WL-Acc. &  Acc. & WL-Acc. &  Acc. & WL-Acc. & \\
    \midrule[1pt]
    Elo & 36.79 & 57.12 & 43.94 & 60.67 & 42.80 & 65.57 & -- \\
    \branch random omission & 36.88 (+0.2\%) & 57.30 (+0.3\%) & 43.77 (-0.3\%) & 60.59 (0.0\%) & 42.73 (-0.2\%)& 65.24 (-0.5\%) & -0.1\% \\[-0.35mm]
    \leaf w/o draw updates & \textbf{38.15}$^\dagger$(+3.7\%) & \textbf{58.32}$^\dagger$(+2.1\%) & \textbf{45.03}$^\dagger$(+2.5\%) & \textbf{61.85}$^\dagger$(+1.9\%) & \textbf{45.07}$^\dagger$(+5.3\%)& \textbf{67.18}$^\dagger$(+2.5\%) & +3.0\% \\
    \midrule[0.1pt]
    Glicko-2 & 40.45 & 61.26 & 46.95 & 65.34 & 46.88 & 69.61 & -- \\
    \branch random omission & 40.41 (0.0\%) & 61.35 (+0.1\%) & 46.93 (0.0\%) & 65.17 (-0.3\%) & 46.86 (0.0\%) & 69.47 (-0.2\%) & -0.1\%\\[-0.35mm]
    \leaf w/o draw updates & \textbf{40.87}$^\dagger$(+1.0\%) & \textbf{61.85}$^\dagger$(+0.9\%) & \textbf{47.91}$^\dagger$(+2.0\%) & \textbf{65.37} (0.0\%) & \textbf{47.03} (+0.3\%) & \textbf{69.74} (+0.1\%) & +0.7\%\\
    \midrule[0.1pt]
    Bradley--Terry & 40.44 & 60.85 & 46.28 & 64.61 & 46.96 & 69.53 & -- \\
    \branch random omission & 40.44 (0.0\%) & 60.58 (-0.4\%) & 46.23 (-0.1\%) & 64.62 (0.0\%) & 46.95 (0.0\%) & 69.58 (0.0\%) & -0.1\% \\[-0.35mm]
    \leaf w/o draw updates & \textbf{40.98}$^\dagger$(+1.3\%) & \textbf{61.30}$^\dagger$(+0.7\%) & \textbf{47.29}$^\dagger$(+2.2\%) & \textbf{65.06}$^\dagger$(+0.6\%) & \textbf{47.46}$^\dagger$(+1.1\%) & \textbf{69.88}$^\dagger$(+0.5\%) & +1.1\% \\
    \midrule[0.1pt]
    TrueSkill & 40.81 & 61.52 & 46.86 & 64.69 & 47.17 & \textbf{69.95} & -- \\
    \branch random omission & 40.83 (0.0\%) & 61.61 (+0.1\%) & 46.85 (0.0\%) & 65.07 (+0.6\%) & 47.20 (0.0\%) & 69.60 (-0.5\%) & 0.0\%\\[-0.35mm]
    \leaf w/o draw updates & \textbf{41.04}$^\dagger$(+0.6\%) & \textbf{62.01}$^\dagger$(+0.7\%) & \textbf{46.90} (0.0\%) & \textbf{65.37}$^\dagger$(+1.1\%) & \textbf{47.45} (+0.6\%) & 69.74 (-0.3\%) & +0.5\%\\
    \bottomrule[1pt]
    \end{tabular}
    \caption{Prequential battle outcome prediction accuracy under various experimental treatments, where ``Acc.'' denotes the overall accuracy and ``WL-Acc.'' the win--loss accuracy if we disallow draws. The relative changes of each ablation with respect to the baselines are in parentheses, and $\Delta\%$ is the global average. Best results are bolded. $^\dagger$One-sided statistical significance at the 95\% level ($p<0.05$) according to McNemar's test~\cite{mcnemar1947note}.}
    \label{tab:main-results}
\end{table*}
\newpage
\section{Experiments}

We selected three open datasets of real-world LLM battles curated from Chatbot Arena: LMArena, SearchArena, and VisionArena.
LMArena~\cite{tang2025explorer} consists of 106K battles from users chatting with 55 text-only LLMs, ranging from LLaMA 3.1-405B~\cite{touvron2024llama3} to GPT-4o.
SearchArena~\cite{miroyan2025search} comprises 24K battles of 13 LLM-driven agents for information access, such as GPT-4o-search.
VisionArena~\cite{chou2025visionarena} has 30K public battles among 17 vision--language models (VLMs), e.g., LLaVA~\cite{liu2023visual}.
Roughly 30--40\% of each dataset are draws, with the remainder split evenly between wins and losses.
%; see Appendix~\ref{X} for detailed statistics.

To evaluate rating systems, we followed \citet{herbrich2006trueskill} and measured prequential battle prediction accuracy:\ we iterated through the battles chronologically, predicting the outcome from the current ratings before updating them.
Concretely, we computed $\frac{1}{n}\sum_{i=1}^n \mathbb{I}(\hat{y}_i = y_i)$, where the prediction $\hat{y}_i \in L$ depends only on $r^{(a_i)}_i$ and $r^{(b_i)}_i$ (and any state at $i$).
TrueSkill predicts draws naturally using the draw margin $\varepsilon$; for Elo, Glicko-2, and Bradley--Terry, we introduced a draw margin $\varepsilon$ in the decision rule:
\begin{equation}
    \hat{y} = \begin{cases}
        0 & \text{if~} E_{b_i} - E_{a_i} > \varepsilon,\\
        \frac{1}{2} &\text{if~} \left| E_{a_i} - E_{b_i} \right| \leq \varepsilon,\\
        1 &\text{if~} E_{a_i} - E_{b_i} > \varepsilon,\\
    \end{cases}
\end{equation}
which can be tuned like any other hyperparameter.

\subsection{Draw Update Ablation Study}

\parheaderfirst{Setup}
We first evaluated the impact of omitting rating updates for draws.
For each dataset, we set aside the first 5\% as the calibration set and the remaining 95\% as the validation set.
We tuned the draw margin $\varepsilon$ on the calibration set, sweeping it in the interval $[0.05, 0.45]$ with a step size of $0.05$.
We then used the best $\varepsilon$ from \textit{including} draw updates for all experiments within each method--dataset combination, as separately tuning it for ignoring draw updates may lead to unfair bias in favor of ignoring updates.
As a baseline, to remove fewer updates as a potential confounder, we also omitted both draws and win--loss updates randomly at a rate equal to the number of draws in each dataset.
% See Appendix~\ref{X} for more details.

\vspace{-0.07mm}
\parheader{Results}
As shown in Table~\ref{tab:main-results}, ignoring draw updates improves outcome prediction accuracy by a relative 0.5--3.0\% on average for all four rating systems, with median overall and win--loss accuracy improvements of 1.2\% and 0.7\%, respectively.
These gains are statistically significant in 18 of 23 cases.
The effect on Elo is most prominent (+3.0\%), followed by Bradley--Terry (+1.1\%), Glicko-2 (+0.7\%) and TrueSkill (+0.5\%), possibly because Elo does not model uncertainty.
With its net-zero change, the random omission ablations also demonstrate that the effect cannot be explained by merely using less data.
Performance-wise, Glicko-2, Bradley--Terry, and TrueSkill are evenly matched, with a median overall accuracy range of 0.42 absolute points (see VisionArena's 47.03--47.46).
This contrasts with Elo, which lags behind the other systems by a median 3.6 points.

A foreseeable concern is that disregarding draw updates may benefit win--loss accuracy but hurt draw prediction accuracy, while still increasing aggregate accuracy.
To address this, we sweep $\varepsilon$ and plot the resulting win--loss-to-draw accuracy curves.
As Figure~\ref{fig:draw-wl-acc-curves} in Appendix~\ref{sec:further-ablation} confirms, ignoring draws improves draw prediction accuracy at all operating points, i.e., it is Pareto-better.

\subsection{Draw Semantics Study}

\parheaderfirst{Setup}
To assess the effect of query difficulty and subjectivity on draws, we sampled 3,000 battles from LMArena and labeled the query's difficulty and subjectivity on a scale from 0--5 using GPT-4.1.
Then, we binned all the outcomes by rating and computed the risk ratio (RR) of observing a draw versus a win or a loss.
RRs above 1.0 represent a higher likelihood of draws and below 1.0 the opposite.
For all 106K battles, we also collected the absolute difference in the model ratings and whether a draw occurred, then binned them by the difference percentile and computed the RR.
\begin{figure}[t]
    \centering
    \includegraphics[width=0.48\columnwidth]{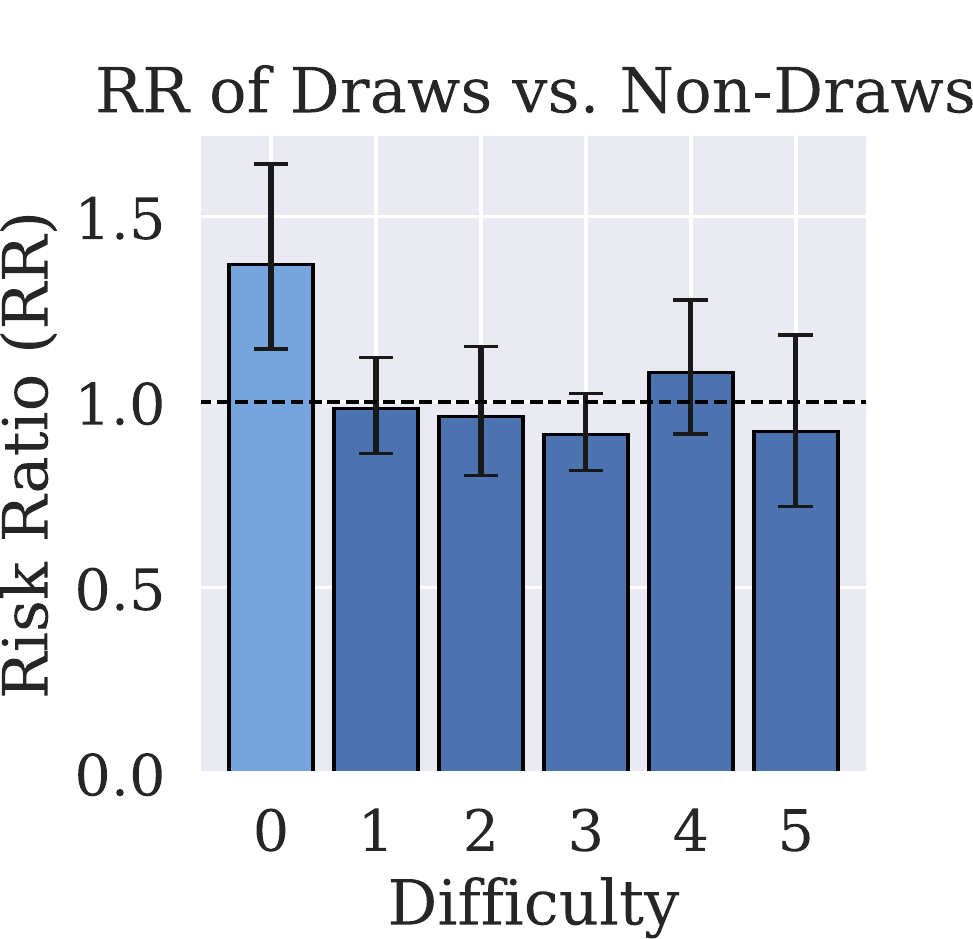}
    \includegraphics[width=0.48\columnwidth]{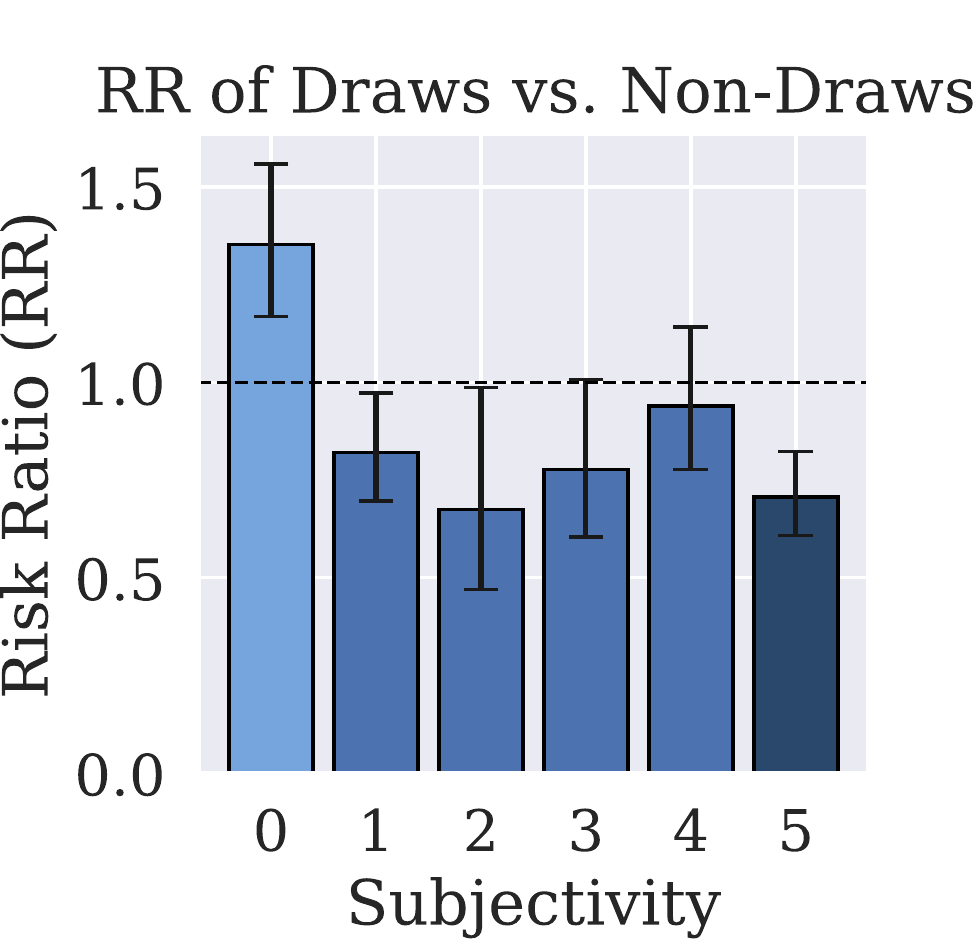}
    \caption{The risk ratio of observing a draw compared to a win or a loss, binned by difficulty and subjectivity. Error bars denote 95\% confidence intervals.}
    \label{fig:difficulty-subjectivity-rrs}
\end{figure}

\begin{figure}[t]
    \centering
    \vspace{1.2mm}
    \includegraphics[width=0.889\columnwidth]{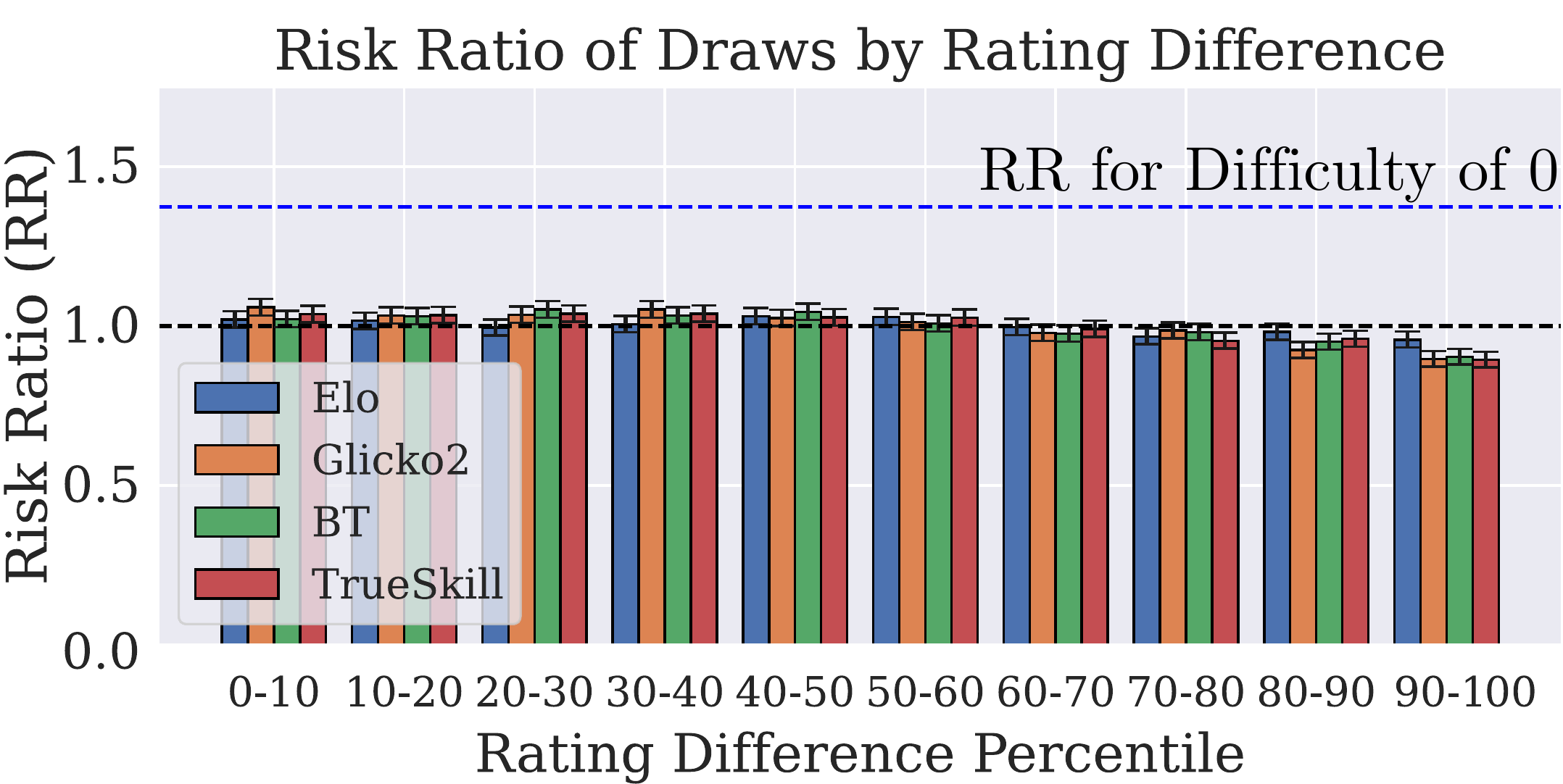}
    \caption{The risk ratio of observing a draw as a function of the absolute difference in model ratings, binned by percentile range.}
    \label{fig:rating-diff-rrs}
\end{figure}

\parheader{Results}
Figure~\ref{fig:difficulty-subjectivity-rrs} shows that draws are indeed more likely for queries with difficulty and subjectivity ratings of 0, which reach respective RRs of 1.37 and 1.35.
This likely follows from very easy queries being broadly answerable by any LLM and highly objective queries having an exact match.
Other ratings are not significantly different from an RR of 1.0, except for highly subjective queries rated as a 5 more likely to result in a win or a loss, possibly due to creative tasks eliciting stronger feedback.
Next, Figure~\ref{fig:rating-diff-rrs} presents the draw RR as a binned function of the rating difference.
If lower percentiles have high RRs, then that suggests rating closeness is predictive of draws.
This was \textit{not} the case, since all RRs are close to 1.0 until the 90--100th percentiles, which only slightly differs at RRs of 0.89--0.96, further affirming our central hypotheses.

\section{Related Work}
Although Chatbot Arena popularized anonymous head-to-head battles for LLMs~\cite{chiang2024chatbot}, ordinal comparisons of LLM responses originated with InstructGPT~\cite{ouyang2022training}, the direct predecessor to ChatGPT.
Recent work has probed pitfalls of pairwise judging, such as position bias~\cite{shi2024judging}, test contamination~\cite{white2024livebench}, and misalignment with real-world utility~\cite{miller2025evaluating}.
Large-scale benchmarks such as BIG-Bench similarly emphasize broad coverage and systematic evaluation across a diverse array of tasks \cite{srivastava2023beyond}.

Other studies critically analyze the robustness of arena-style evaluation, focusing on the Elo rating system.
\citet{boubdir2023elouncovered} show that Elo can violate desirable axioms such as reliability and transitivity; \citet{liuelo} address these issues with ``am-ELO,'' a maximum-likelihood reformulation that jointly models annotator reliability.
\citet{wu2025style} find that single Elo scores overweight stylistic fluency over factual correctness, motivating the Multi-Elo Rating System (MERS).
Our work complements this body of literature by questioning draw semantics in arena evaluation.

\section{Conclusions}
In this work, we questioned the standard assumption that draws in arena-style LLM evaluation indicate model parity. 
Across three real-world datasets, ignoring draw updates improved outcome prediction accuracy by 1--3\%, despite draws comprising 30--40\% of battles. 
Our analysis further showed that draws are disproportionately associated with very easy and highly objective queries (risk ratios of 1.37 and 1.35), suggesting they stem more from query properties.
For future rating systems, we recommend reconsidering what draws mean and to explicitly account for query properties.

\balance
\bibliography{anthology}

\begin{thebibliography}{21}
\providecommand{\natexlab}[1]{#1}

\bibitem[{Boubdir et~al.(2023)Boubdir, Kim, Ermis, Hooker, and Fadaee}]{boubdir2023elouncovered}
Meriem Boubdir, Edward Kim, Beyza Ermis, Sara Hooker, and Marzieh Fadaee. 2023.
\newblock Elo uncovered: Robustness and best practices in language model evaluation.
\newblock In \emph{Proceedings of the Third Workshop on Natural Language Generation, Evaluation, and Metrics (GEM)}.

\bibitem[{Bradley and Terry(1952)}]{bradley1952rank}
Ralph~A. Bradley and Milton~E. Terry. 1952.
\newblock Rank analysis of incomplete block designs.
\newblock \emph{Biometrika}.

\bibitem[{Chiang et~al.(2023)Chiang, Li, Gonzalez, and Stoica}]{chiang2023chatbot}
Wei-Lin Chiang, Tim Li, Joseph~E. Gonzalez, and Ion Stoica. 2023.
\newblock Chatbot {A}rena: New models \& {E}lo system update.

\bibitem[{Chiang et~al.(2024)Chiang, Zheng, Sheng, Angelopoulos, Li, Li, Zhu, Zhang, Jordan, Gonzalez et~al.}]{chiang2024chatbot}
Wei-Lin Chiang, Lianmin Zheng, Ying Sheng, Anastasios~N. Angelopoulos, Tianle Li, Dacheng Li, Banghua Zhu, Hao Zhang, Michael Jordan, Joseph~E. Gonzalez, et~al. 2024.
\newblock Chatbot {A}rena: An open platform for evaluating {LLMs} by human preference.
\newblock In \emph{ICML}.

\bibitem[{Chou et~al.(2025)Chou, Dunlap, Mashita, Mandal, Darrell, Stoica, Gonzalez, and Chiang}]{chou2025visionarena}
Christopher Chou, Lisa Dunlap, Koki Mashita, Krishna Mandal, Trevor Darrell, Ion Stoica, Joseph~E. Gonzalez, and Wei-Lin Chiang. 2025.
\newblock {VisionArena}: 230{K} real world user-{VLM} conversations with preference labels.
\newblock In \emph{CVPR}.

\bibitem[{Elo(1978)}]{elo1978rating}
Arpad~E. Elo. 1978.
\newblock \emph{The Rating of Chessplayers, Past and Present}.

\bibitem[{Glickman(2012)}]{glickman2012example}
Mark~E. Glickman. 2012.
\newblock Example of the {G}licko-2 system.
\newblock \emph{Boston University}.

\bibitem[{Glickman and Jones(2024)}]{glickman2024models}
Mark~E. Glickman and Albyn~C. Jones. 2024.
\newblock Models and rating systems for head-to-head competition.
\newblock \emph{Annual Review of Statistics and Its Application}.

\bibitem[{Grattafiori et~al.(2024)Grattafiori, Ainslie, Bhosale, and et~al.}]{touvron2024llama3}
Andrea Grattafiori, Joshua Ainslie, Shruti Bhosale, and et~al. 2024.
\newblock The {LLaMA} 3 herd of models.
\newblock \emph{arXiv:2407.21783}.

\bibitem[{Herbrich et~al.(2006)Herbrich, Minka, and Graepel}]{herbrich2006trueskill}
Ralf Herbrich, Tom Minka, and Thore Graepel. 2006.
\newblock {TrueSkill}: A {B}ayesian skill rating system.
\newblock \emph{NeurIPS}.

\bibitem[{Liu et~al.(2023)Liu, Li, Wu, and Lee}]{liu2023visual}
Haotian Liu, Chunyuan Li, Qingyang Wu, and Yong~Jae Lee. 2023.
\newblock Visual instruction tuning.
\newblock \emph{NeurIPS}.

\bibitem[{Liu et~al.(2025)Liu, Li, Zhuang, Liu, Shen, Ouyang, Cheng, and Wang}]{liuelo}
Zirui Liu, Jiatong Li, Yan Zhuang, Qi~Liu, Shuanghong Shen, Jie Ouyang, Mingyue Cheng, and Shijin Wang. 2025.
\newblock am-{ELO}: A stable framework for arena-based {LLM} evaluation.
\newblock In \emph{ICML}.

\bibitem[{McNemar(1947)}]{mcnemar1947note}
Quinn McNemar. 1947.
\newblock Note on the sampling error of the difference between correlated proportions or percentages.
\newblock \emph{Psychometrika}.

\bibitem[{Miller and Tang(2025)}]{miller2025evaluating}
Justin~K. Miller and Wenjia Tang. 2025.
\newblock Evaluating {LLM} metrics through real-world capabilities.
\newblock \emph{arXiv:2505.08253}.

\bibitem[{Miroyan et~al.(2025)Miroyan, Wu, King, Li, Pan, Hu, Chiang, Angelopoulos, Darrell, Norouzi et~al.}]{miroyan2025search}
Mihran Miroyan, Tsung-Han Wu, Logan King, Tianle Li, Jiayi Pan, Xinyan Hu, Wei-Lin Chiang, Anastasios~N. Angelopoulos, Trevor Darrell, Narges Norouzi, et~al. 2025.
\newblock Search {A}rena: Analyzing search-augmented {LLM}s.
\newblock \emph{arXiv:2506.05334}.

\bibitem[{Ouyang et~al.(2022)Ouyang, Wu, Jiang, Almeida, Wainwright, Mishkin, Zhang, Agarwal, Slama, Ray et~al.}]{ouyang2022training}
Long Ouyang, Jeffrey Wu, Xu~Jiang, Diogo Almeida, Carroll Wainwright, Pamela Mishkin, Chong Zhang, Sandhini Agarwal, Katarina Slama, Alex Ray, et~al. 2022.
\newblock Training language models to follow instructions with human feedback.
\newblock \emph{NeurIPS}.

\bibitem[{Shi et~al.(2024)Shi, Ma, Liang, Diao, Ma, and Vosoughi}]{shi2024judging}
Lin Shi, Chiyu Ma, Wenhua Liang, Xingjian Diao, Weicheng Ma, and Soroush Vosoughi. 2024.
\newblock Judging the judges: A systematic study of position bias in {LLM}-as-a-judge.
\newblock \emph{arXiv:2406.07791}.

\bibitem[{Srivastava et~al.(2023)Srivastava, Rastogi, Rao, Shoeb, Abid, Fisch, Brown, Santoro, Gupta, Garriga-Alonso et~al.}]{srivastava2023beyond}
Aarohi Srivastava, Abhinav Rastogi, Abhishek Rao, Abu~Awal Shoeb, Abubakar Abid, Adam Fisch, Adam~R Brown, Adam Santoro, Aditya Gupta, Adri Garriga-Alonso, et~al. 2023.
\newblock Beyond the imitation game: Quantifying and extrapolating the capabilities of language models.
\newblock \emph{TMLR}.

\bibitem[{Tang et~al.(2025)Tang, Chiang, and Angelopoulos}]{tang2025explorer}
Kelly Tang, Wei-Lin Chiang, and Anastasios~N. Angelopoulos. 2025.
\newblock Arena {E}xplorer: A topic modeling pipeline for {LLM} evals \& analytics.

\bibitem[{White et~al.(2024)White, Dooley, Roberts, Pal, Feuer, Jain, Shwartz-Ziv, Jain, Saifullah, Dey et~al.}]{white2024livebench}
Colin White, Samuel Dooley, Manley Roberts, Arka Pal, Ben Feuer, Siddhartha Jain, Ravid Shwartz-Ziv, Neel Jain, Khalid Saifullah, Sreemanti Dey, et~al. 2024.
\newblock {LiveBench}: A challenging, contamination-limited {LLM} benchmark.
\newblock \emph{arXiv:2406.19314}.

\bibitem[{Wu and Aji(2025)}]{wu2025style}
Minghao Wu and Alham~Fikri Aji. 2025.
\newblock Style over substance: Evaluation biases for large language models.
\newblock In \emph{COLING}.

\end{thebibliography}
\begin{figure*}[ht!]
    \includegraphics[width=0.95\textwidth]{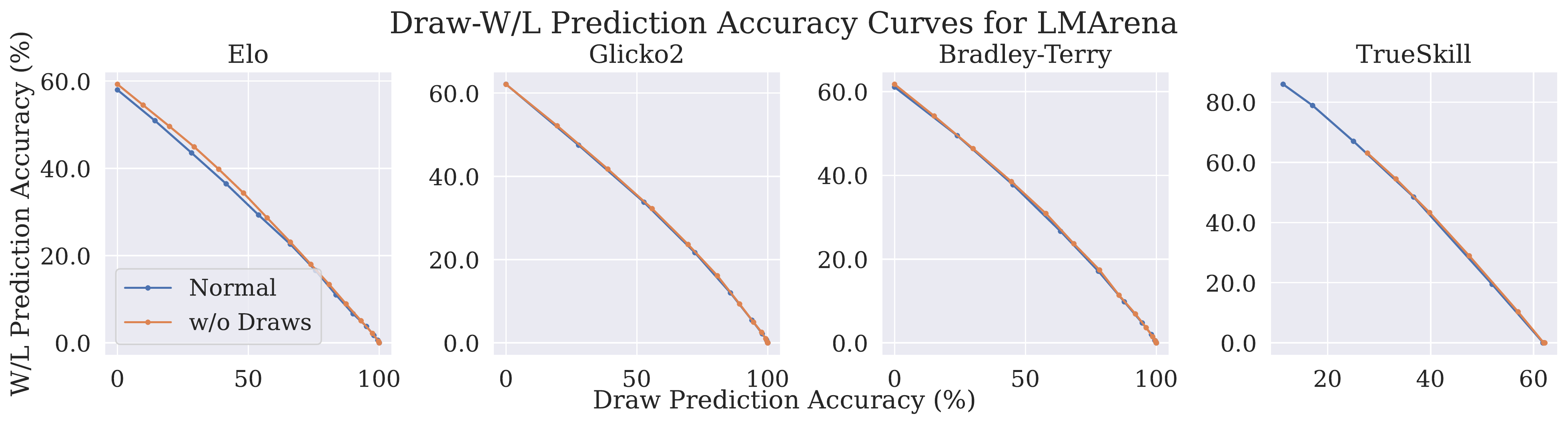}\vspace{2mm}
    \includegraphics[width=0.95\textwidth]{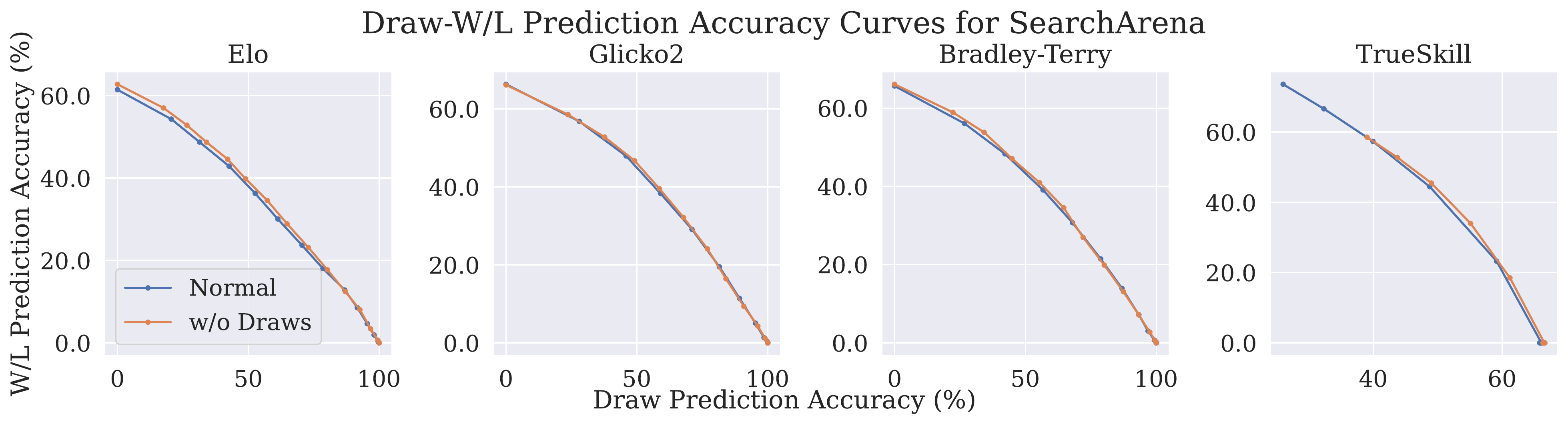}\vspace{2mm}
    \includegraphics[width=0.95\textwidth]{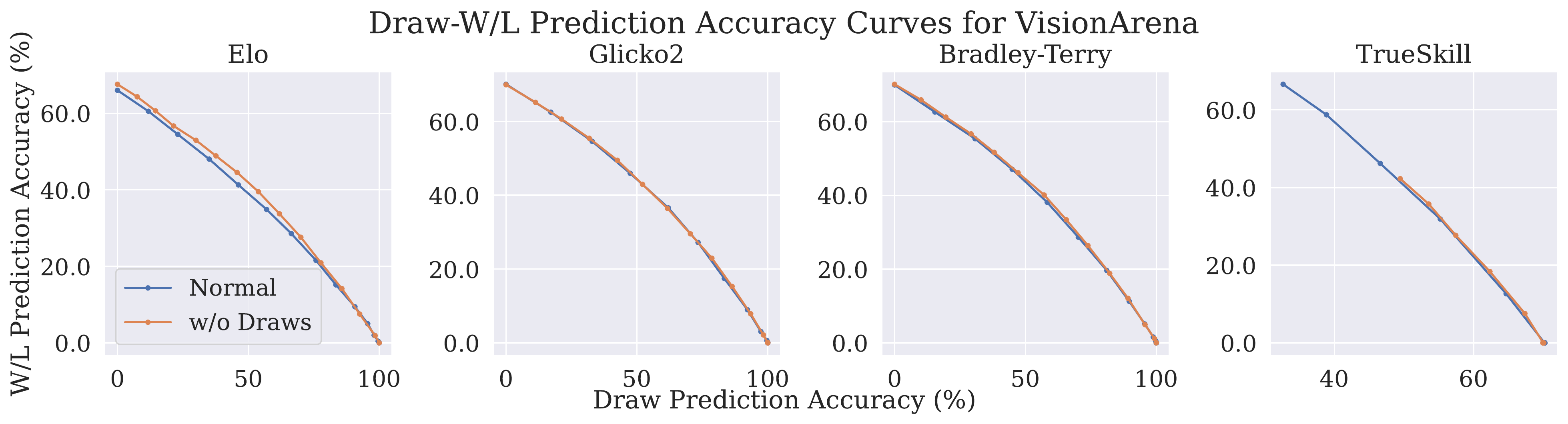}
    \caption{Trade-off curves between draw and win/loss prediction accuracy as we vary the draw margin. Larger draw margins result in better draw prediction accuracy at the expense of win/loss accuracy, and vice versa. Curves with higher maxima and AUC are better.}
    \label{fig:draw-wl-acc-curves}
\end{figure*}
\begin{figure*}[ht!]
\centering
\includegraphics[width=0.85\textwidth]{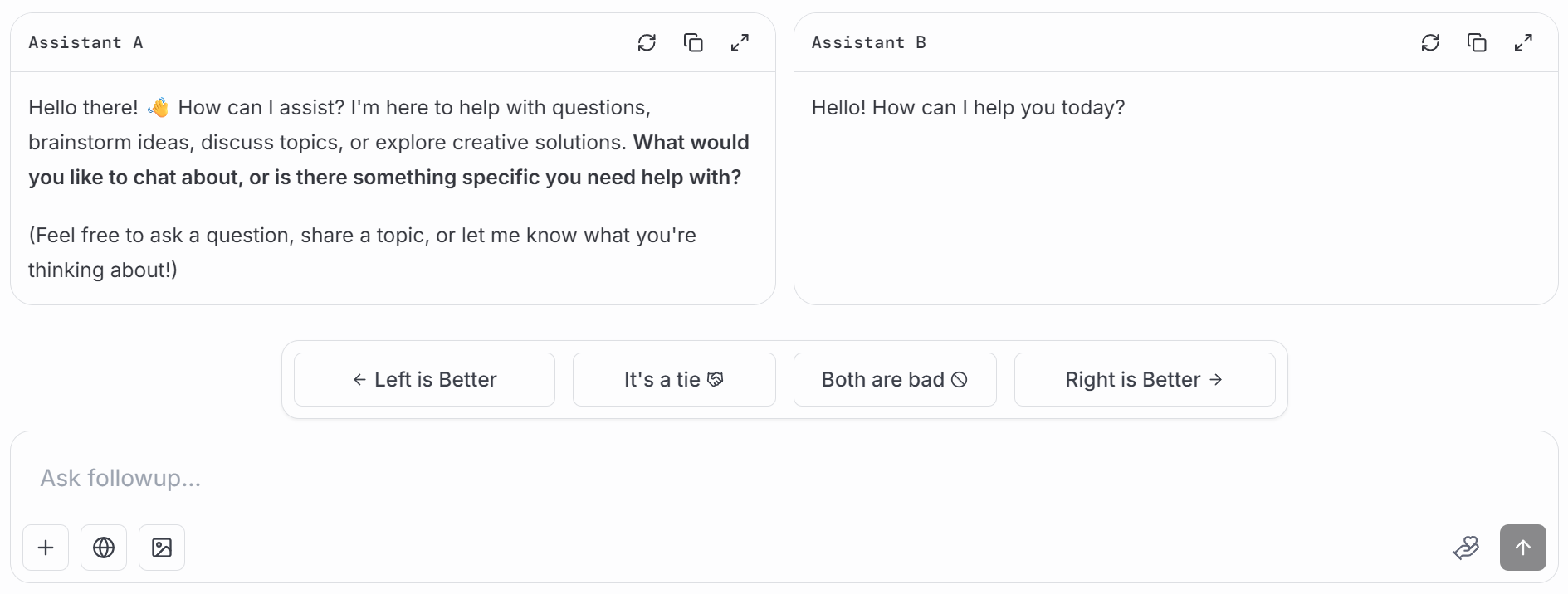}
\caption{An example user interface from \url{https://lmarena.ai}.}
\label{fig:example-ui}
\end{figure*}
\newpage
\nobalance
\appendix
\section{Further Ablation}
\label{sec:further-ablation}
In Figure~\ref{fig:draw-wl-acc-curves}, we vary the draw threshold $\varepsilon$ and plot the trade-off curves between draw and win--loss prediction accuracy.
Ignoring draw updates~(orange line) attains a higher AUC and is Pareto-better than including everything.
\section{Example User Interface}
\label{sec:appendix-ui}
We present an example user interface of arena-style evaluation in Figure~\ref{fig:example-ui}.
Users can choose left is better, right is better, or a draw.

\clearpage
\newpage

\end{document}